\documentclass[conference]{IEEEtran}
\IEEEoverridecommandlockouts

\usepackage{cite}
\usepackage{amsmath,amssymb,amsfonts}
\usepackage{algorithmic}
\usepackage{graphicx}
\usepackage{textcomp}
\usepackage{xcolor}
\def\BibTeX{{\rm B\kern-.05em{\sc i\kern-.025em b}\kern-.08em
    T\kern-.1667em\lower.7ex\hbox{E}\kern-.125emX}}
\usepackage{amsmath,amsfonts}
\usepackage{amssymb,amsthm}
\usepackage[normalem]{ulem}
\usepackage{textcomp}
\usepackage{url}
\usepackage{verbatim}
\usepackage{graphicx}
\usepackage{graphics} 
\usepackage{mathtools}
\usepackage{booktabs}
\usepackage{adjustbox}
\usepackage{stfloats}
\usepackage[breaklinks=true,colorlinks,citecolor=blue]{hyperref}
\usepackage{makecell}
\usepackage{multirow}
\usepackage{balance}
\usepackage{rotating}
\usepackage[linesnumbered,ruled,vlined,commentsnumbered]{algorithm2e}
\usepackage{footnote}
\usepackage[dvipsnames]{xcolor}
\usepackage{tabularray}
\usepackage{pifont}
\usepackage{censor}

\newcommand{\orchid}[1]{{\color{Orchid}{#1}}}

\newcommand{\SEthree}{\mathrm{SE}(3)}

\begin{document}

\title{\texttt{\textbf{\orchid{GeLiFiSH}}}: Recursive LiDAR Keypoint Estimation with Ground and Moving Object Segmentation}

\author{
\thanks{The authors are with the Robotics and AI Group, Department of Computer, Electrical and Space Engineering, Luleå University of Technology, 971 87 Luleå, Sweden. Corresponding Author’s e-mail: niksta@ltu.se}
\thanks{Additional demonstration material, examples and the source code of this paper will be available upon acceptance at our github repo.}
Nikolaos Stathoulopoulos and George Nikolakopoulos}


\maketitle

\begin{abstract}
LiDAR registration and loop closure often rely on sparse keypoints, which must remain repeatable across sessions, viewpoints, and sensors.
To handle differing scan patterns, detectors typically operate on accumulated submaps, which introduces latency and processes each submap independently, so repeated observations are never fused.
Height-based detectors also reference structure to an estimated ground, so its inconsistency across sessions propagates into the keypoints.
Moreover, dynamic objects leave ghost trails that can be mistaken for static structure.
We present GeLiFiSH (Geometric LiDAR Filtering with Sequential Hypothesis testing), a learning-free filter that recursively estimates persistent keypoints at scan rate.
A height-structure detector runs on a rolling voxel map, and detections initialize tracks with a position, covariance, and existence probability, accepted or rejected by a sequential probability ratio test.
Ground segmentation provides a sensor-agnostic height reference, and moving object segmentation removes dynamic points before map insertion.
On SemanticKITTI, HeLiMOS, and HeLiPR, spanning five LiDARs and multiple sessions, GeLiFiSH attains the highest average keypoint repeatability among the compared detectors and the second-highest cross-session registration recall from keypoints alone, behind ISS which extracts four times as many keypoints.
With one parameter set and no sensor-height prior, both ground and moving object segmentation modules outperform their baselines on average.
\end{abstract}

\begin{IEEEkeywords}
LiDAR keypoints, heterogeneous LiDARs, ground segmentation, moving object segmentation
\end{IEEEkeywords}

\section{Introduction}~\label{sec:introduction}
Lifelong mapping~\cite{gil_ephemerality_2025}, multi-robot map merging~\cite{stathoulopoulos_frame_2024,kim2026commerge}, and relocalization in prior maps~\cite{3dsurveyAV} all require data association between LiDAR scans captured at different times, from different viewpoints, and increasingly by heterogeneous sensors, as fleets combine spinning, solid-state, and FMCW LiDARs~\cite{minwoo2024helipr}.
Registration, place recognition, and loop closure~\cite{loopclosures,Zhang2024autolipr} commonly reduce each scan to a sparse set of keypoints, which keeps correspondence search tractable at map scale~\cite{yuan2024btc,gupta2024efficiently,stathoulopoulos2026inlier}.
This assumes repeatable keypoints, re-detected on the same structure across sessions regardless of the sensor.
\begin{figure}
    \centering
    \includegraphics[width=1.0\columnwidth]{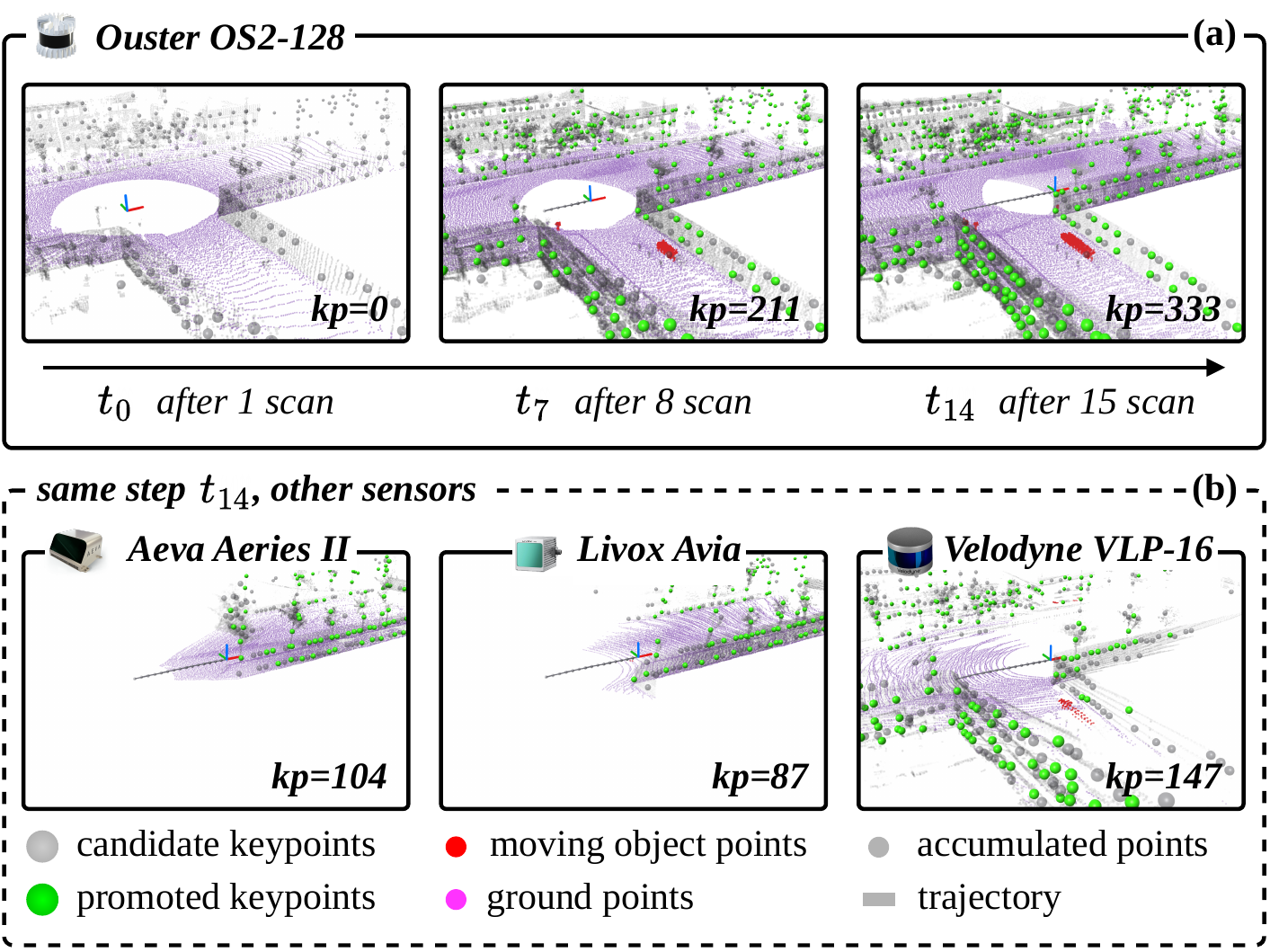}
    \setlength{\abovecaptionskip}{-0.4cm}
    \caption{GeLiFiSH on HeLiPR~\cite{minwoo2024helipr} from a cold start, with the same parameters for every sensor. 
    (a)~Ouster OS2-128 after 1, 8, and 15 scans. Each scan adds its ground (purple) and moving (red) labels and updates the rolling map (gray). Detections enter as candidate keypoints (gray spheres) and are promoted (green) once the sequential test has collected enough evidence, so none is promoted after the first scan. 
    (b)~The same step $t_{14}$ for the other three LiDARs, which differ in density, field of view, and scan pattern.}
    \label{fig:concept}
    \vspace{-0.7cm}
\end{figure}

Heterogeneous LiDARs differ in beam count, field of view, and scan pattern, so a 16-beam spinning sensor and a solid-state sensor with a non-repetitive pattern sample the same scene at very different densities.
Detectors designed for place recognition and loop closure therefore operate on submaps accumulated over several scans to normalize the point density~\cite{yuan2024btc,gupta2024efficiently,stathoulopoulos2026inlier}.
This introduces latency, and since each submap is processed independently, repeated detections of the same structure are never fused.
Dynamic objects crossing the submap further leave ghost trails that resemble static structure, and for height-based detectors, any inconsistency of the estimated ground across sessions biases the height reference and, with it, the keypoints.

We present GeLiFiSH (Geometric LiDAR Filtering with Sequential Hypothesis testing), a learning-free filter that recursively estimates persistent keypoints at scan rate.
It applies the InLiER detector~\cite{stathoulopoulos2026inlier} unchanged to a rolling voxel map that normalizes the point density across sensors, and models each keypoint as a track with a position, a covariance, and an existence probability.
Since consecutive maps share most of their points, a detection counts as a hit only if the newest scan resampled its cell, and a track counts as a miss only if beams of the newest scan traverse it, which keeps the evidence independent across scans.
A sequential probability ratio test (SPRT) accumulates these hits and misses, accepts a track once the evidence suffices, and rejects it once the structure disappears, e.g., when a parked car leaves.
Two learning-free segmentation modules precede the detector and additionally provide per-point labels for downstream use.
Ground segmentation propagates a coarse ground plane between scans as its prior, in place of a sensor-height parameter, and thereby provides a sensor-agnostic height reference.
Moving object segmentation labels points that occupy space previously observed as free and removes them before map insertion, so dynamic objects leave no ghost trails.

In summary, our contributions are:
\begin{itemize}
    \item A recursive keypoint estimator that maintains detections on a rolling map as tracks and confirms them through an SPRT, with hits gated on novel observations and misses on free-space evidence.
    \item A learning-free ground segmentation that propagates its ground plane between scans in place of a sensor-height parameter, and serves as the common height reference of the pipeline.
    \item A learning-free moving object segmentation that fuses visibility checks in recent range images with a long-horizon free-space voxel memory, and expands seed points into object-level segments.
\end{itemize}
We evaluate keypoint repeatability and keypoint-only registration recall on HeLiPR across sessions and four LiDARs against five detectors, and both segmentation modules on SemanticKITTI and HeLiMOS.

\section{Related Work}~\label{sec:related-work}
We review ground segmentation, moving object segmentation, and keypoint extraction, focusing on the methods we evaluate against.

\subsection{Ground Segmentation}
Early approaches segment the ground per scan with handcrafted geometric rules, such as line fitting in angular sectors~\cite{fastgs} or local convexity~\cite{3dlidarseg}.
Patchwork~\cite{lim2021patchwork} handles range-dependent sparsity with a concentric zone model, fits a plane per bin, and rejects false positives through ground likelihood estimation.
GroundGrid~\cite{steinke2024groundgrid} estimates the terrain on an ego-centric 2D elevation grid, and B-TMS~\cite{btms} models traversable terrain in a Bayesian framework, on both single scans and accumulated maps.
Patchwork++~\cite{lee_patchwork_2022} extends Patchwork with thresholds adapted to the elevation and flatness of previously segmented ground, reflected noise removal, region-wise vertical plane fitting, and temporal ground revert against partial under-segmentation.
It requires the sensor height as a parameter and was designed for dense 64-beam spinning LiDARs.
GeLiFiSH instead propagates the ground plane of the previous scan as its prior, so a single parameter set serves sparse, dense, and solid-state sensors alike.

\subsection{Moving Object Segmentation}
MOS separates points on moving objects from the static scene, and existing methods either learn motion cues from labeled data or infer them from free space observed by earlier scans.
4DMOS~\cite{mersch20224dmos} applies sparse 4D convolutions to a short window of pose-aligned scans and fuses overlapping predictions over a receding horizon, refining each label as later scans arrive.
MapMOS~\cite{mersch2023mapmos} replaces the window with a local map, segments moving points in both the scan and the map, and fuses the predictions over time in a volumetric belief.
Both are trained on the Velodyne HDL-64E of SemanticKITTI~\cite{behley2019iccv}.
The learning-free HMM-MOS~\cite{bhandari2026hmmmos} raycasts each scan into a voxel map, updates per-voxel occupancy with a hidden Markov model, and labels voxels whose occupancy changes as moving after spatiotemporal convolution filtering.
GeLiFiSH, also learning-free, fuses range-image visibility with a long-horizon free-space voxel memory and grows the resulting seeds into object-level segments above the tracked ground.

\subsection{Keypoint Extraction}
Keypoint detectors select a sparse set of salient points, which should be repeatable across viewpoints and sensors.
ISS~\cite{zhong2009iss}, among the most widely used handcrafted detectors, selects points whose local scatter matrix has well-separated eigenvalues and applies non-maximum suppression on the smallest one.
RSKDD-Net~\cite{lu2020rskdd} learns a detector and a descriptor jointly from randomly sampled points and is trained on KITTI~\cite{behley2019iccv}.
Detectors built for place recognition and loop closure typically operate on submaps accumulated over several scans to normalize the point density.
BTC~\cite{yuan2024btc} extracts keypoints from plane boundaries and corners of the submap and encodes them in binary and triangle descriptors.
MapClosures~\cite{gupta2024efficiently} projects the submap to a BEV density image and detects ORB features on it.
InLiER~\cite{stathoulopoulos2026inlier}, learning-free and designed for heterogeneous LiDARs, slices the submap by height above its estimated ground and detects local maxima of the vertical extent on a BEV grid in each slice.
GeLiFiSH runs the InLiER detector on a rolling map updated at scan rate and maintains each detection as a track whose existence an SPRT confirms, so the keypoints persist across scans.

\section{Methodology}~\label{sec:methodology}
GeLiFiSH processes a stream of LiDAR scans, each registered into a gravity-aligned map frame by an odometry pose $\mathbf{T}_t=(\mathbf{R}_t,\mathbf{x}_t)\in\SEthree$, with rotation $\mathbf{R}_t$ and sensor position $\mathbf{x}_t\in\mathbb{R}^3$.
For every scan, it outputs per-point ground and moving labels and a set of persistent keypoints.
All modules operate on the same preprocessed scan $\bar{\mathcal{P}}_t$, obtained by voxelizing the raw points $\mathcal{P}_t$ at $v_\text{in}$ and cropping them to a horizontal range $[R_{\min},R_{\max}]$, which removes ego-vehicle returns and overly sparse far-range returns.
The modules form a feed-forward chain, each consuming only the outputs of its predecessors.
Ground segmentation (GS, Sec.~\ref{subsec:gs}) operates per scan in the sensor frame and propagates a coarse ground plane $\mathbf{g}_t$ between scans, which, transformed by $\mathbf{T}_t$, serves as the common ground reference of the pipeline.
Moving object segmentation (MOS, Sec.~\ref{subsec:mos}) uses this plane to separate objects from their supporting surface, and labels as moving the points that contradict free space observed by earlier scans.
Recursive keypoint estimation (RKE, Sec.~\ref{subsec:rke}) accumulates the points labeled static in a rolling local map, references the height slices of the InLiER~\cite{stathoulopoulos2026inlier} detector to $\mathbf{g}_t$, and maintains the detections as tracks that an SPRT accepts or rejects, with the newest range image from MOS as free-space evidence.

\subsection{Ground Segmentation (GS)}~\label{subsec:gs}
Given a scan $\bar{\mathcal{P}}_t=\{\mathbf{p}_i\in\mathbb{R}^3\}$ in the sensor frame, we seek the ground points $\mathcal{G}_t \subset \bar{\mathcal{P}}_t$. 
The segmentation is pose- and map-free, and its only state is a coarse ground plane $\mathbf{g}_t=(z_t,\boldsymbol{\gamma}_t)$, i.e.\ the surface:
\begin{equation}
    z = z_t + \boldsymbol{\gamma}_t^\top\mathbf{x}_\text{xy},
    \label{eq:plane}
\end{equation}
with height $z_t$ below the sensor and gradient $\boldsymbol{\gamma}_t\in\mathbb{R}^2$ at horizontal position $\mathbf{x}_\text{xy}\in\mathbb{R}^2$. 
Since a single plane cannot follow curbs, shoulders or slope changes up to $R_{\max}$, we refine it with local patches grown outward from it, as in Patchwork++~\cite{lee_patchwork_2022}, and label each point against the local surface beneath it. 

\subsubsection{Plane initialization}
For the first scan, no prior plane exists to define the band through residuals. 
We therefore seed $z_t$ with RANSAC on the points within $r_g=R_{\max}/4$ that lie less than $b=1$\,m above their $5^{\text{th}}$ height percentile, followed by three PCA refits with halving thresholds ($b$, $b/2$, $b/4$). 

\subsubsection{Plane tracking}
Points within $[R_{\min},R_{\max}]$ and within $[-2,\,22]$\,m height of the previous plane $\mathbf{g}_{t-1}$ are retained, which follows slopes and sensor pitch while discarding below-ground multipath returns and high structure.
They are voxelized at $v_g=0.5$\,m, which normalizes the range-dependent density and reduces the cost of the fit.
Only voxels within $r_g$, where the ground is densest and the prior most reliable, enter the fit, with their heights expressed as residuals to $\mathbf{g}_{t-1}$ so that a slope or pitch does not clip the band.
The voxels less than $b$ above the $5^{\text{th}}$ residual percentile form the band $\mathcal{B}_0$, with centers $(\mathbf{x}_{\text{xy},i},z_i)$.
Using a percentile rather than the minimum keeps a few below-ground multipath returns from lowering the band, similarly to the reflected noise removal of Patchwork++.
The plane $\hat{\mathbf{g}}_t=(\hat{z}_t,\hat{\boldsymbol{\gamma}}_t)$ is the least-squares solution:
\begin{equation}
    \hat{\mathbf{g}}_t^{(k)}=\operatorname*{arg\,min}_{z,\boldsymbol{\gamma}}\sum_{i\in\mathcal{B}_k}\left(z_i-z-\boldsymbol{\gamma}^\top\mathbf{x}_{\text{xy},i}\right)^2,
    \label{eq:ls}
\end{equation}
which is refitted twice, $k=1,2$, on the voxels within $r_g$ whose vertical distance to the previous fit is below $b/2^k$:
\begin{equation}
    \mathcal{B}_k=\left\{i:\ \left|z_i-\hat{z}^{(k-1)}-\hat{\boldsymbol{\gamma}}^{(k-1)\top}\mathbf{x}_{\text{xy},i}\right|<b/2^k\right\},
\end{equation}
with $\hat{\mathbf{g}}_t=\hat{\mathbf{g}}_t^{(2)}$, where the shrinking tolerance rejects car bodies and wall bases that the band still admits. The gradient is taken from the fit, while the height is averaged with the prior:
\begin{equation}
    z_t = \tfrac{1}{2}\left(z_{t-1}+\hat{z}_t\right),
\end{equation}
which damps single-scan disturbances, e.g. a vehicle in view. 

\subsubsection{Ground patches}
The ground is divided into $c_s \times c_s$ columns, with $c_s= 2\,v_g$, each of which may receive a local plane:
\begin{equation}
    \pi_c(\mathbf{p}) = z_c + \boldsymbol{\gamma}_c^\top(\mathbf{p}_{xy}-\mathbf{c}_c),
\end{equation}
with height $z_c$ at the column center $\mathbf{c}_c$ and gradient $\boldsymbol{\gamma}_c$.
The lowest occupied voxel of a column is its ground candidate, and the column is skipped when the two voxel layers directly above it are occupied, since such gap-free structure marks the foot of a wall, fence or trunk and would tilt the fitted plane.
The plane of column $c$ is fitted by PCA to the candidates of its $3\times3$ neighborhood up to $R_{\max}/2$, and of its $5\times5$ neighborhood beyond, where the ground is sparser. As the candidates are the lowest voxels, this plane is biased low, so it is refitted twice on the scan points of the block's non-skipped columns that lie within $v_g/2$, and then $v_g/4$, of the plane of their own column.
A patch is usable if its column was not skipped and it is upright, with a tilt below $45^\circ$ ($\lVert\boldsymbol{\gamma}_c\rVert\le 1$), and flat, with a root-mean-square (RMS) point-to-plane distance below 0.1\,m.
Usable columns are linked to their eight neighbors when their planes differ by at most $\delta_{\text{step}}=0.25$\,m in height at the shared border, and a usable column within $r_g$ whose center height lies within $\delta_s=0.3$\,m of $\mathbf{g}_t$ is a seed. All columns connected to a seed are accepted, and the rest have no patch. Growing from seeds near the sensor rejects flat but disconnected surfaces such as roofs and car tops, while the step limit admits curbs and slope changes but not walls.

\subsubsection{Point labels}
The height of a point above the ground is measured against the accepted patch of its column $c$, $h_i=z_i-\pi_c(\mathbf{p}_i)$, or against $\mathbf{g}_t$ if the column has none, $h_i=z_i-z_t-\boldsymbol{\gamma}_t^\top\mathbf{p}_{i,xy}$. Points with $h_i<\delta=0.2$\,m pass the height test, which has no lower bound, so that points below the surface, e.g.\ in potholes, remain ground.
Since skipped columns have no patch, a wall foot can still pass the test against the plane, so we detect the wall feet $\mathcal{W}_t$ per point at a finer resolution. Around each point with $h_i<\delta$ within $r_g$, a $10\times10$\,cm column of the scan is cut into 5\,cm layers from $-0.5$ to $+1.5$\,m about the ground, and the point belongs to $\mathcal{W}_t$ if at least 2 of the 6 layers directly above its own are occupied, as a wall rises without a gap while a car body or a bush starts higher.
The ground points are those that pass the height test and are not wall feet:
\begin{equation}
    \mathcal{G}_t=\left\{\mathbf{p}_i\in\bar{\mathcal{P}}_t:\ h_i<\delta\right\}\setminus\mathcal{W}_t.
    \label{eq:ground}
\end{equation}

\subsection{Moving Object Segmentation (MOS)}~\label{subsec:mos}
Given a scan $\bar{\mathcal{P}}_t$ and its pose $\mathbf{T}_t$, we seek the moving points $\mathcal{M}_t\subset\bar{\mathcal{P}}_t$.
The cue is visibility, as a static point is hit or occluded in every earlier view that could observe it, whereas a moving object occupies space that earlier beams traversed.
Here the points $\mathbf{p}_i$ are transformed by $\mathbf{T}_t$, and $\tilde h_i$ is their height above $\mathbf{g}_t$, transformed likewise, which separates an object from the surface it stands on.
Free-space evidence is kept in two representations, a ring of range images of the recent scans and a voxel memory that accumulates it over a longer horizon.
Both yield seed points, which region growing expands into object-level segments.
Each scan is labeled twice from the same state, online from the scans up to it, and again $D=10$ scans later, when the subsequent scans also vote.

\subsubsection{Visibility votes}
The last $K=30$ scans are stored as range images centered at their sensor positions $\mathbf{x}_j$, where $I_j(\mathbf{b})$ is the range of the nearest return in the $0.3^\circ\times0.3^\circ$ direction bin $\mathbf{b}$.
The bin matches the $0.2^\circ$ to $0.4^\circ$ beam spacing of the dense sensors, so few bins remain empty, and at the reach $R_\text{MOS}=50$\,m of the voxel memory it spans $0.26$\,m, about one memory voxel, so a pedestrian still covers about two bins.
Storing the nearest return is conservative, since only a beam whose return lies beyond a point proves its spot empty.
Two steps prevent an image from claiming unobserved free space.
First, each bin takes the minimum over its $3\times3$ neighborhood, since from a new viewpoint the silhouette of an object falls in a bin that the earlier scan filled with the background.
Second, for sensors with few beams, e.g.\ 16 rings $2^\circ$ apart, an empty elevation run of up to $4^\circ$ between two returns of an azimuth column takes the nearer of the two ranges.
Empty bins carry no information.

Let $r_{ij}=\lVert\mathbf{p}_i-\mathbf{x}_j\rVert$ be the range of $\mathbf{p}_i$ from $\mathbf{x}_j$, and $I_{ij}=I_j(\mathbf{b}_i)$ the range that image $j$ stores in its direction $\mathbf{b}_i$.
Image $j$ votes free if its beam passed more than $\Delta r=v_g$ beyond the point, and reaches the point if its beam passed or hit the same surface, giving the free votes $n_i$ and reaching views $a_i$:
\begin{equation}
    n_i=\sum_{j}\big[I_{ij}>r_{ij}+\Delta r\big],\quad
    a_i=\sum_{j}\big[I_{ij}>r_{ij}-\Delta r\big],
    \label{eq:votes}
\end{equation}
over the $K-1$ previous scans with a non-empty bin in that direction, where $[\cdot]$ is the Iverson bracket.
The ratio $n_i/a_i$ is thus the fraction of views that could observe the spot and found it empty, and occluded views ($I_{ij}\le r_{ij}-\Delta r$) enter neither sum, so they cannot dilute it.
A point is a seed if $n_i\ge n_{\min}=2$, $n_i\ge\eta\,a_i$ with $\eta=0.55$, and $\tilde h_i\ge h_{\min}=\delta_s$, where the height gate, the seed tolerance of GS, keeps the ground static, and the fractional threshold accounts for the differing coverage of the sensors.

\subsubsection{Voxel memory}
Over $K$ scans, a sparse sensor returns from a given spot in only a few images, so a mover there collects too few votes.
The voxel memory therefore accumulates the evidence per voxel of size $v_m=v_g/2$ over all its observations, and drops the columns beyond $R_\text{MOS}$.
Each voxel $q$ holds a free count $\phi_q$ and a reach count $\psi_q$.
A scan observes $q$ as hit if one of its points falls in it, and as free if the nearest return within its angular footprint, dilated by $\varepsilon=0.15$\,m laterally and vertically, lies more than $\Delta r$ beyond its center.
The dilation gives the free test a 3D tolerance, similar to the distance field of HMM-MOS~\cite{bhandari2026hmmmos}, since with a range margin alone a wall at grazing incidence turns centimeters of lateral offset into meters of range and appears free between its hits.
Occluded, partially occluded and near-surface voxels are left unchanged, and an observed voxel is updated by $\phi_q\leftarrow\lambda\,\phi_q+[\text{free}]$ and $\psi_q\leftarrow\lambda\,\psi_q+1$.
The decay $\lambda=0.97$ applies per observation of the voxel, so a rarely observed voxel retains its evidence.
A point reads the counts of its voxel before its own scan updates them and is a seed under the image rule, $\phi_q\ge n_{\min}$, $\phi_q\ge\eta\,\psi_q$ and $\tilde h_i\ge h_{\min}$, and the seeds of both representations form $\mathcal{Z}_t$.

\subsubsection{Region growing}
Seeds are sparse on movers, whose points next to a surface may never be observed free, and spurious on foliage and facades, so the decision is made per object.
The scan points with $\tilde h_i \ge h_{\min}$ are binned on a BEV grid with cell size $v_m$ and clustered into 8-connected components $\mathcal{C}_k$, and a component is moving if its seeds make up at least a fraction $\kappa=0.3$ of its points.
With $\bar{\mathcal{C}}_k\supseteq\mathcal{C}_k$ the scan points in the cells of $\mathcal{C}_k$, the moving points are:
\begin{equation}
    \mathcal{M}_t=\left\{\mathbf{p}_i\in\bar{\mathcal{C}}_k:\ \lvert\mathcal{C}_k\cap\mathcal{Z}_t\rvert\ge\kappa\lvert\mathcal{C}_k\rvert\right\}\setminus\mathcal{G}_t,
    \label{eq:moving}
\end{equation}
where removing the ground points of~\eqref{eq:ground} keeps the road beneath the object static.

\subsubsection{Delayed labels}
A receding mover is self-occluded in the images before $t$, so its spot is observed free only after it has left, which the online label cannot capture.
The label of scan $t$ is therefore reissued at $t+D$ with the same seeding and growing, with the sums of~\eqref{eq:votes} running over the $K-1$ scans before $t$ and the $D$ scans after it, and the voxel counts extended by the free observations $F_q$ and hits $H_q$ since $t$ to $\phi_q+F_q$ and $\psi_q+F_q+H_q$.
An object that has left its voxel adds free observations, whereas a parked vehicle keeps adding hits and remains static.

\subsection{Recursive Keypoint Estimation (RKE)}~\label{subsec:rke}
Given a scan $\bar{\mathcal{P}}_t$ and its pose $\mathbf{T}_t$, we seek the persistent keypoints $\mathcal{K}_t=\{\mathbf{k}_i\}$.
The InLiER detector~\cite{stathoulopoulos2026inlier} runs on a rolling buffer updated at every scan, which normalizes the point density and re-detects the same structure from many viewpoints as the platform moves.
Each keypoint is a track $\mathbf{k}_i=(\boldsymbol{\mu}_i,\boldsymbol{\Sigma}_i,\boldsymbol{\Sigma}_i^\text{shape},\ell_{i,t})$ with a mean position $\boldsymbol{\mu}_i\in\mathbb{R}^3$ in the map frame, a position covariance $\boldsymbol{\Sigma}_i$, a shape covariance $\boldsymbol{\Sigma}_i^\text{shape}$ and an existence log-odds $\ell_{i,t}$, and is reported once a sequential test accepts it as persistent.

\subsubsection{Buffer and detections}
The buffer $\mathcal{B}_t$ accumulates the static points $\bar{\mathcal{P}}_t\setminus\mathcal{M}_t$ of~\eqref{eq:moving}, so that movers leave no ghost trails, and is voxelized at $v_b=c_s/2$, half the detector cell size $c_s$, with one centroid per voxel.
InLiER runs on $\mathcal{B}_t$ unchanged, cutting it into $N_h$ height slices above $\mathbf{g}_t$, binning each slice on a BEV grid of cell size $c_s$ that stores the vertical extent of its points, and returning the maxima of $3\times3$ non-maximum suppression (NMS) as detections $\mathbf{d}_j$ at the centroid of their points.
The grid is fixed in the map frame, so that static structure falls in the same cell at every scan.

\subsubsection{Evidence}
The sequential test requires evidence that is independent between scans, so each scan contributes a novelty-gated hit, a seen-through miss, or no evidence to a track.

\textit{Novelty:} 
Consecutive buffers share most of their points, so their detections are near-duplicates.
A detection is novel only if the newest scan resampled at least a fraction $\rho=0.3$ of the occupied voxels of its cell, which ties it to the current viewpoint and, as a fraction, transfers across sensor densities.

\textit{Hit:} 
Novel detections are associated greedily, nearest first, with the tracks of their slice within a gate of less than one cell, and each match is a hit.
An unmatched novel detection initializes a track with its first hit, and non-novel detections are discarded.

\textit{Seen-through miss:} 
A missing detection is uninformative, since the keypoint may lie outside the field of view or behind an occluder while the buffer retains its points.
Yet a track on a departed object, e.g.\ a parked car that drove off, receives no novel detections, and without negative evidence it would persist where nothing stands.
A track without a novel match is therefore missed only if the newest scan saw through it, i.e.\ if no non-ground return in the angular footprint of its cell in $I_t$ lies nearer than the keypoint plus a margin $\Delta r_k=c_s$, and at least one return lies beyond it.
Ground returns are ignored as blockers, since a beam that reaches the road short of the keypoint passes beneath it, whereas road beyond the keypoint shows that the beam went past, and empty bins are skipped.

\subsubsection{Sequential test}
With the binary existence state $e_i\in\{0,1\}$ of track $i$, $\mathcal{E}=\{e_i\}$, and the buffers $\mathcal{B}_{1:t}$ and range images $I_{1:t}$ up to scan $t$, we assume independent tracks, so that the belief factorizes into:
\begin{equation}
    p(\mathcal{E}\mid \mathcal{B}_{1:t},I_{1:t})=\prod_i p(e_i\mid \mathcal{B}_{1:t},I_{1:t}).
    \label{eq:belief}
\end{equation}
Bayes' rule yields the recursive binary Bayes filter~\cite{thrun2005probabilistic}, which in the log-odds $\ell(x)=\log\frac{p(x)}{1-p(x)}$ reads:
\begin{equation}
\begin{split}
    \ell(e_i\mid \mathcal{B}_{1:t},I_{1:t})=\ &\ell(e_i\mid \mathcal{B}_{1:t-1},I_{1:t-1})\\
    &+\ell(e_i\mid \mathcal{B}_t,I_t)-\ell(e_i).
\end{split}
    \label{eq:sprt}
\end{equation}
Here $\ell_{i,t}=\ell(e_i\mid \mathcal{B}_{1:t},I_{1:t})$ is stored in the track and clipped to $[-\ell_{\max},\ell_{\max}]$, and the prior $p(e_i)=0.5$ gives $\ell(e_i)=0$.
The inverse sensor model $\ell(e_i\mid \mathcal{B}_t,I_t)$ is $\ell_\text{hit}>0$ for a hit, $\ell_\text{miss}<0$ for a seen-through miss, and zero otherwise, so that $\ell_{i,t}$ is the log-likelihood ratio of a sequential probability ratio test with thresholds $\pm\tau$.
A track becomes persistent when $\ell_{i,t}>\tau$ with at least $m_{\min}=4$ hits, is dropped when $\ell_{i,t}<-\tau$, and reverts to a candidate when $\ell_{i,t}$ falls back to $\tau$, which the clipping at $\ell_{\max}$ slows down, so that a few misses from an occluder cannot demote a well-supported track.

\subsubsection{Estimate}
The position $\boldsymbol{\mu}_i$ is the mean of the detections $\mathbf{d}_{i,1},\dots,\mathbf{d}_{i,m_i}$ of its $m_i$ hits, which averages the jitter of the detection within its cell over the viewpoints, and its covariance is the standard error of this mean:
\begin{equation}
    \boldsymbol{\Sigma}_i=\frac{1}{m_i^2}\Big(\boldsymbol{\Psi}_{0}+\sum_{h=1}^{m_i}(\mathbf{d}_{i,h}-\boldsymbol{\mu}_i)(\mathbf{d}_{i,h}-\boldsymbol{\mu}_i)^\top\Big),
    \label{eq:cov}
\end{equation}
where the novelty gate makes the hits independent, so that $\boldsymbol{\Sigma}_i$ shrinks with $m_i$.
The prior $\boldsymbol{\Psi}_{0}=\tfrac{1}{12}\operatorname{diag}(c_s^2,c_s^2,h_{\delta z}^2)$, with $h_{\delta z}$ the slice thickness, is the covariance of a position uniform within one cell and slice, and counts as one detection, so that $\boldsymbol{\Sigma}_i$ is defined from the first hit.
Each keypoint additionally carries the shape covariance $\boldsymbol{\Sigma}_i^\text{shape}$ of the buffer points around $\boldsymbol{\mu}_i$, which describes the local geometry as in InLiER.

\section{Experimental Evaluation}~\label{sec:results}
All methods receive identical scans, cropped to a horizontal range of $[R_{\min},R_{\max}]=[5,80]$\,m and voxelized at $v_\text{in}=0.1$\,m.
GeLiFiSH uses a single parameter set across all datasets and sensors, with the values given in Sec.~\ref{sec:methodology} and the following RKE parameters.
The detector retains the InLiER defaults, a cell size of $c_s=1$\,m and $N_h=10$ height slices over $[0,20]$\,m, i.e.\ a slice thickness of $h_{\delta z}=2$\,m.
The sequential test uses $\ell_\text{hit}=0.85$, $\ell_\text{miss}=-0.3$, $\tau=2$, and $\ell_{\max}=8$.
Runtimes are measured on a single core of an Intel i9-14900K, and the learned methods run on an NVIDIA RTX 4090.
Each module is evaluated on its respective task.

\subsection{Ground Segmentation}
We compare ground segmentation against Patchwork++ \cite{lee_patchwork_2022}, the most widely adopted state-of-the-art method for this task.
On SemanticKITTI~\cite{behley2019iccv} 00--10, we follow the evaluation protocol of Patchwork++, with road, parking, sidewalk, other-ground, lane-marking, and terrain as ground and vegetation excluded from scoring.
Precision, recall, and F1 score are computed per scan, averaged per sequence, and then averaged across sequences.
To assess cross-sensor consistency, we use HeLiMOS~\cite{helimos}, recorded on the KAIST05 sequence of HeLiPR~\cite{minwoo2024helipr} with an Ouster OS2-128, an Aeva Aeries II, a Livox Avia, and a Velodyne VLP-16 mounted on the same vehicle.
As HeLiMOS provides no ground labels, the labels of each sensor are compared against those of the denser Ouster at the same timestamp.
Both label sets are transformed into the world frame with the ground-truth poses and voxelized at 0.2\,m, each voxel taking the majority label of its points.
Voxels containing a moving point are discarded, and the voxels observed by both sensors are compared over approximately 12.5k scans per sensor.
We report the IoU of the ground class and the fraction of voxels with identical labels (agreement).
On both datasets, Patchwork++ runs with its default parameters and is given the mounting height of each sensor, a prior that GeLiFiSH does not require.

On SemanticKITTI (Table~\ref{tab:ground_kitti}), GeLiFiSH achieves an F1 score of 96.9\,\% against 96.3\,\% for Patchwork++, at a comparable runtime (4.2 vs.\ 4.0\,ms) and without the sensor height.
The ablation isolates the two components of GS.
The patches increase recall from 94.7 to 97.2\,\% by recovering the ground that deviates from the plane at curbs and slope changes, at a cost of 0.4 points in precision.
The prior leaves F1 unchanged and reduces the runtime from 10.4 to 4.2\,ms, as it replaces the RANSAC initialization that every scan would otherwise need.

Across the HeLiMOS sensors (Table~\ref{tab:ground_helipr}), the labels of GeLiFiSH are more consistent with those of the Ouster on all three sensors, with a mean IoU of 80.5\,\% against 70.4\,\% and an agreement of 95.4\,\% against 91.9\,\%, even though Patchwork++ is given each mounting height.
The largest gap occurs on the VLP-16, with 69.7\,\% against 51.3\,\% IoU.
Patchwork++ partitions the scan with a fixed concentric zone model of zones, rings, and sectors, whose defaults are tuned for the 64-beam sensor of SemanticKITTI, and labels any bin with fewer than ten points entirely as non-ground.
Sensors with fewer beams or a narrower field of view leave more bins below this threshold, most notably the 16-beam VLP-16.
GeLiFiSH operates on metric voxels and columns, independent of the beam pattern.
The voxelization before the fit normalizes the point density, the patches are fitted on larger neighborhoods where the ground is sparser, and the plane propagated from the previous scan supports sparse scans.
Since the ground is the reference of the height slices, this consistency propagates to the keypoints across sensors.
\begin{table}[t]
\centering
\footnotesize
\caption{Ground segmentation on SemanticKITTI 00--10. Patchwork++'s evaluation protocol. Precision (P), Recall (R), F1: mean $\pm$ std.\ over sequences; time: mean $\pm$ std.\ per scan on one core.}
\label{tab:ground_kitti}
\setlength{\tabcolsep}{2pt}
\begin{tabular*}{\columnwidth}{@{\extracolsep{\fill}}lcccccc@{}}
\toprule
 & & & \multicolumn{4}{c}{Mean 00--10} \\
\cmidrule(lr){4-7}
Method & Prior & Patches & P [\%] & R [\%] & F1 [\%] & Time [ms] \\
\midrule
Patchwork++ & -- & \checkmark & 95.6\,{\scriptsize$\pm$2.4} & 97.2\,{\scriptsize$\pm$1.0} & 96.3\,{\scriptsize$\pm$1.3} & 4.0\,{\scriptsize$\pm$0.6} \\
\midrule
\multirow{4}{*}{GeLiFiSH} & -- & -- & 96.9\,{\scriptsize$\pm$1.3} & 95.1\,{\scriptsize$\pm$3.3} & 95.9\,{\scriptsize$\pm$2.0} & 8.4\,{\scriptsize$\pm$1.5} \\
 & \checkmark & -- & \textbf{97.1}\,{\scriptsize$\pm$1.2} & 94.7\,{\scriptsize$\pm$3.1} & 95.8\,{\scriptsize$\pm$1.9} & \textbf{2.3}\,{\scriptsize$\pm$0.4} \\
 & -- & \checkmark & 96.6\,{\scriptsize$\pm$1.4} & \textbf{97.4}\,{\scriptsize$\pm$1.4} & 96.9\,{\scriptsize$\pm$1.2} & 10.4\,{\scriptsize$\pm$2.1} \\
 & \checkmark & \checkmark & 96.7\,{\scriptsize$\pm$1.3} & 97.2\,{\scriptsize$\pm$1.5} & \textbf{96.9}\,{\scriptsize$\pm$1.2} & 4.2\,{\scriptsize$\pm$1.0} \\
\bottomrule
\end{tabular*}
\end{table}
\begin{table}[t]
\centering
\footnotesize
\caption{Cross-sensor ground consistency on HeLiMOS KAIST05: each sensor's ground labels against the Ouster OS2-128's.}
\label{tab:ground_helipr}
\setlength{\tabcolsep}{2pt}
\begin{tabular*}{\columnwidth}{@{\extracolsep{\fill}}llccccc@{}}
\toprule
Method & Metric & Ouster & Aeva & Avia & VLP-16 & Mean \\
\midrule
\multirow{2}{*}{Patchwork++} & IoU [\%] & -- & 75.6 & 84.1 & 51.3 & 70.4 \\
 & Agree. [\%] & -- & 94.7 & 93.5 & 87.6 & 91.9 \\
\midrule
\multirow{2}{*}{GeLiFiSH} & IoU [\%] & -- & \textbf{80.1} & \textbf{91.8} & \textbf{69.7} & \textbf{80.5} \\
 & Agree. [\%] & -- & \textbf{95.8} & \textbf{96.7} & \textbf{93.8} & \textbf{95.4} \\
\bottomrule
\end{tabular*}
\end{table}

\subsection{Moving Object Segmentation}
We compare moving object segmentation against the learning-based MapMOS \cite{mersch2023mapmos} and 4DMOS \cite{mersch20224dmos} and the learning-free HMM-MOS \cite{bhandari2026hmmmos}.
On HeLiMOS, we use the test split.
On SemanticKITTI, we evaluate all scans of 00--10, except for the learned methods, whose published weights were trained on 00--07, 09, and 10 and which are therefore evaluated on the validation sequence 08 only.
MapMOS and 4DMOS use these weights with their default configurations, except that MapMOS uses its HeLiMOS inference configuration on HeLiMOS, and HMM-MOS uses its HeLiMOS configuration on both datasets.
MapMOS and 4DMOS estimate their poses with their own KISS-ICP odometry~\cite{vizzo2023ral}, whereas GeLiFiSH and HMM-MOS use the ground-truth poses.
In addition to the online labels, GeLiFiSH, MapMOS, and 4DMOS provide labels delayed by ten scans, obtained from the votes of the subsequent scans, the volumetric belief, and the receding-horizon fusion, respectively.
We report the IoU of the moving class, along with precision and recall, all computed from the true positives, false positives, and false negatives accumulated over all scans.

Table~\ref{tab:mos} lists the IoU of the moving class.
In the online setting, GeLiFiSH attains the highest mean IoU over the four HeLiMOS sensors, 56.7\,\% against 54.9\,\% for MapMOS, 48.0\,\% for 4DMOS, and 44.7\,\% for HMM-MOS.
MapMOS performs best on the Ouster, the Aeva, and the Avia, with a margin of 16.6 points on the dense Ouster.
The advantage of GeLiFiSH stems from the VLP-16, where the learned methods drop to 13.2 and 19.9\,\% while GeLiFiSH retains 43.1\,\%.
There, MapMOS remains precise (98.0\,\%) but reaches a recall of only 13.2\,\%.
With a delay of ten scans, GeLiFiSH improves to 55.9\,\% on the VLP-16 and to 61.6\,\% on average, against 58.7\,\% for MapMOS.
On SemanticKITTI 08, recorded with the sensor on which the learned methods were trained, they clearly lead, with 76.7\,\% for MapMOS against 38.4\,\% for GeLiFiSH online.
Over 00--10, GeLiFiSH outperforms HMM-MOS, with 55.4\,\% against 43.8\,\%.
GeLiFiSH labels a HeLiMOS scan in 18.0 (Avia) to 48.8\,ms (Ouster) on a single CPU core, including GS and the delayed labels, and in 27.9\,ms on average, four to twelve times faster than HMM-MOS on one core (333\,ms) and than MapMOS and 4DMOS on the GPU (239 and 121\,ms, including their KISS-ICP odometry).
The learned methods are thus the most accurate on the denser sensors and on their training sensor, but degrade sharply on the sparse VLP-16, indicating limited generalization across sensors.
GeLiFiSH trades part of that accuracy for a single parameter set that transfers to the VLP-16, at a fraction of their computational cost on a single CPU core.

\begin{table}[b]
\centering
\footnotesize
\caption{Moving object segmentation: IoU of the moving class [\%] on HeLiMOS and SemanticKITTI. Online: a scan is labeled from the scans up to it; delayed: labeled 10 scans late. Best per column within a latency group in bold.}
\label{tab:mos}
\setlength{\tabcolsep}{2pt}
\begin{tabular*}{\columnwidth}{@{\extracolsep{\fill}}llcccccccc@{}}
\toprule
 & & & \multicolumn{5}{c}{HeLiMOS} & \multicolumn{2}{c}{SemanticKITTI} \\
\cmidrule(lr){4-8}\cmidrule(l){9-10}
 & Method & ML & Ouster & Aeva & Avia & VLP-16 & Mean & 08 & 00--10 \\
\midrule
\multirow{4}{*}{\rotatebox{90}{Online}} & MapMOS & \checkmark & \textbf{81.8} & \textbf{66.5} & \textbf{58.1} & 13.2 & 54.9 & \textbf{76.7} & -- \\
 & 4DMOS & \checkmark & 60.7 & 61.5 & 49.8 & 19.9 & 48.0 & 67.0 & -- \\
 & HMM-MOS & -- & 53.9 & 54.8 & 51.5 & 18.6 & 44.7 & 29.3 & 43.8 \\
 & GeLiFiSH & -- & 65.2 & 64.0 & 54.6 & \textbf{43.1} & \textbf{56.7} & 38.4 & \textbf{55.4} \\
\midrule
\multirow{4}{*}{\rotatebox{90}{Delayed}} & MapMOS & \checkmark & \textbf{83.0} & \textbf{68.1} & \textbf{62.1} & 21.7 & 58.7 & \textbf{76.9} & -- \\
 & 4DMOS & \checkmark & 70.0 & 66.2 & 57.8 & 32.0 & 56.5 & 70.3 & -- \\
 & HMM-MOS & -- & -- & -- & -- & -- & -- & -- & -- \\
 & GeLiFiSH & -- & 66.0 & 63.9 & 60.8 & \textbf{55.9} & \textbf{61.6} & 47.1 & \textbf{62.1} \\
\bottomrule
\end{tabular*}
\end{table}

\subsection{Keypoint Estimation}
Finally, we compare the keypoints against the handcrafted ISS~\cite{zhong2009iss}, the learned RSKDD-Net~\cite{lu2020rskdd}, the keypoints that BTC~\cite{yuan2024btc} and MapClosures~\cite{gupta2024efficiently} use for place recognition, and the standalone InLiER detector~\cite{stathoulopoulos2026inlier}, which GeLiFiSH runs within RKE.
We use three HeLiPR~\cite{minwoo2024helipr} routes, each traversed in two separate sessions, Roundabout01$\rightarrow$03, Town01$\rightarrow$03, and Bridge02$\rightarrow$03.
The Ouster of the earlier session serves as the reference and is matched against each of the four sensors of the later session.
RSKDD-Net uses its KITTI weights and returns 512 keypoints per window regardless of the point density.
BTC runs with its published configuration, MapClosures with its ORB detector on the BEV density image, and InLiER with its own ground estimate.
At each keyframe, the input is a window of 10 scans, or 20 for the sparser Avia and VLP-16, accumulated in the keyframe frame with the ground-truth poses.
The baselines detect on this accumulated submap, whereas GeLiFiSH is initialized empty, processes the scans of the window sequentially, and outputs its persistent keypoints after the last scan.

We first evaluate the repeatability of the keypoints across sessions and sensors.
For each sensor of the later session, every keyframe is paired with the later scan closest to an XY offset of 1, 3, 5, and 10\,m among those with a heading within $10^\circ$ of it, and, to cover viewpoint changes, with the closest later scan within 5\,m in each of four bins of heading difference.
For a pair of windows $X$ and $Y$, let $\mathcal{K}_X$ and $\mathcal{K}_Y$ be the keypoints that have a point of the other window within 1\,m under the ground-truth relative pose, which restricts both sets to the overlap.
Two keypoints match one-to-one if they are mutual nearest neighbors within $\epsilon=1$\,m, and the repeatability $r$ of the pair is the matched fraction of the sparser set.
This combines the distance criterion of 3D detector evaluation~\cite{tombari2013performance}, which normalizes by the keypoints of one set, with the normalization of image-based repeatability~\cite{mikolajczyk2005comparison}, which matches regions by their overlap.
Since denser sets match more often by chance, we also compute the repeatability $r_0$ of equally many random points, drawn from each overlap downsampled to 0.5\,m voxels so that they are uniform over the observed surface, and correct for it over all pairs as:
\begin{equation}
    r=\frac{M}{\min\left(\lvert\mathcal{K}_X\rvert,\lvert\mathcal{K}_Y\rvert\right)},\qquad\kappa=\frac{R-R_0}{1-R_0},
    \label{eq:kappa}
\end{equation}
where $M$ is the number of matches, and $R$, $R_0$ are the means of $r$ and $r_0$ over the pairs.
The chance-corrected repeatability $\kappa$ thus normalizes the gain over chance by its maximum, such that random keypoints yield $\kappa=0$ and $\kappa=100\,\%$ indicates that every keypoint of the sparser set is matched.

We then assess whether the keypoints alone suffice to register the two windows of a pair.
For the pairs at the four offsets, the two keypoint sets are aligned by point-to-point ICP~\cite{small_gicp} initialized at the identity, with mutual nearest-neighbor correspondences within a gate that shrinks to 0.5\,m and eight iterations per gate.
A pair is successfully registered if the relative translation error (RTE) is below 0.5\,m and the relative rotation error (RRE) below $2^\circ$, and we report the the fraction of successfully registered pairs (registration recall).

\begin{figure}[t]
    \centering
    \includegraphics[width=1.0\columnwidth]{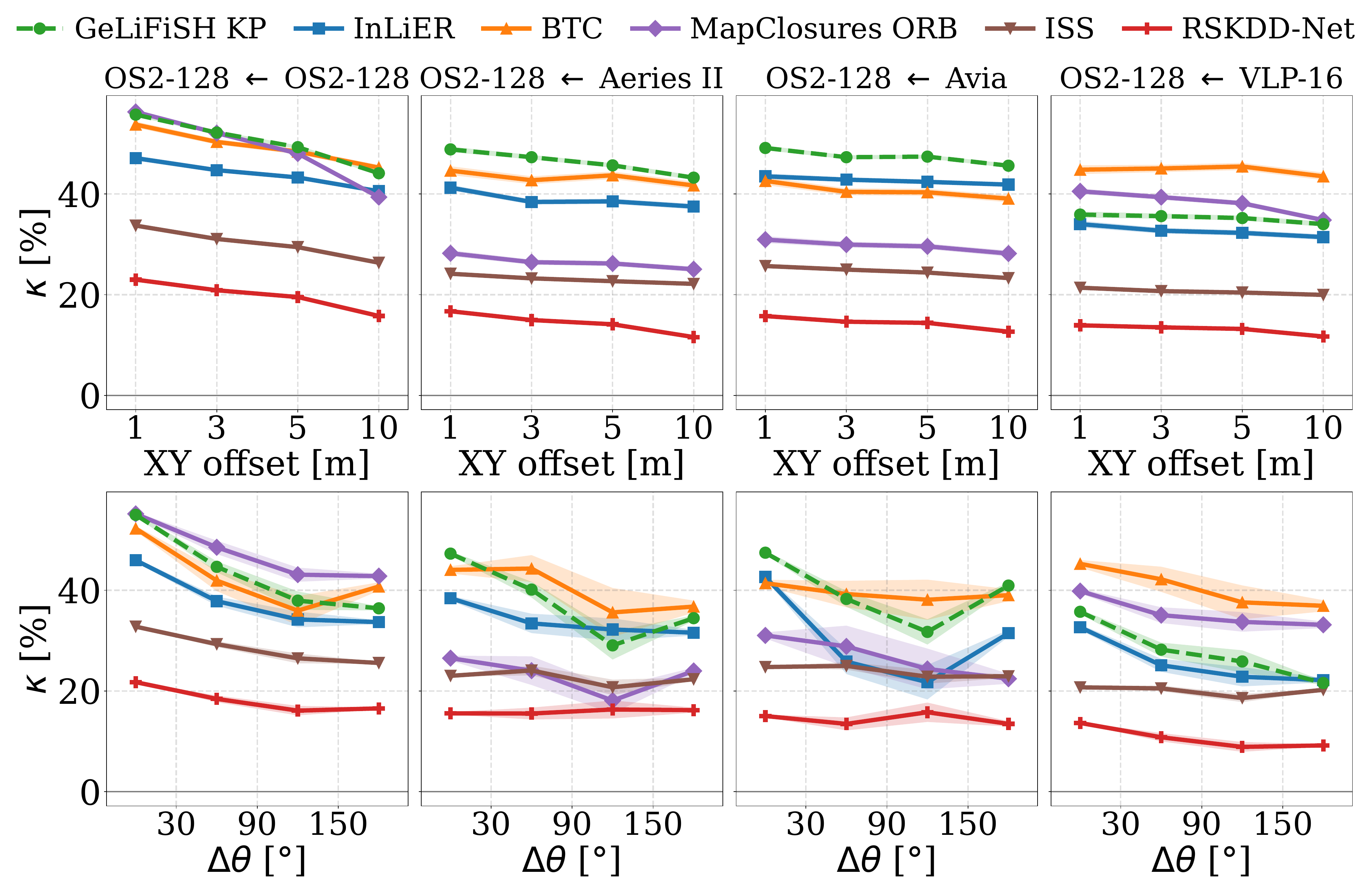}
    \caption{Chance-corrected repeatability $\kappa$ at $\epsilon=1$\,m on HeLiPR, with the Ouster OS2-128 of the earlier session as reference against each sensor of the later session. The top row is binned by the XY offset between the two poses (headings within $10^\circ$), the bottom row by their heading difference (offset within 5\,m). Shading is $\pm1$ standard deviation over a bootstrap of the keyframes.}
    \label{fig:repeatability}
\end{figure}

Averaged over the four sensors and offsets (Fig.~\ref{fig:repeatability}), GeLiFiSH achieves $\kappa=44.8\,\%$, followed by BTC (44.5\,\%), InLiER (39.5\,\%), MapClosures (35.8\,\%), ISS (24.6\,\%), and RSKDD-Net (15.4\,\%).
GeLiFiSH exceeds InLiER at every offset and on every sensor.
Since both use the same detector, the gain is attributable to the components that GeLiFiSH adds around it, namely the rolling map, the tracks, and the ground and moving labels.
Across sensors, GeLiFiSH is the most repeatable on the Aeva and the Avia at every offset.
On the Ouster, MapClosures, GeLiFiSH, and BTC lie within 2.5 points of each other up to 5\,m, whereas on the VLP-16, BTC (44.8\,\% at 1\,m) and MapClosures (40.5\,\%) outperform GeLiFiSH (35.9\,\%).
As $\kappa$ is corrected for chance at the keypoint count of each method, it allows a fair comparison of sets of very different sizes.
In the overlap at 1\,m, BTC retains only 22 to 87 keypoints per window and ISS 418 to 1512, against 101 to 392 for GeLiFiSH.
ISS yields dense sets, such that random points at its counts already reach a repeatability of 31 to 33\,\%, and its $\kappa$ remains between 21 and 34\,\%.
RSKDD-Net, with 512 keypoints for every window, is the least repeatable on every sensor.

\begin{figure}[t]
    \centering
    \includegraphics[width=1.0\columnwidth]{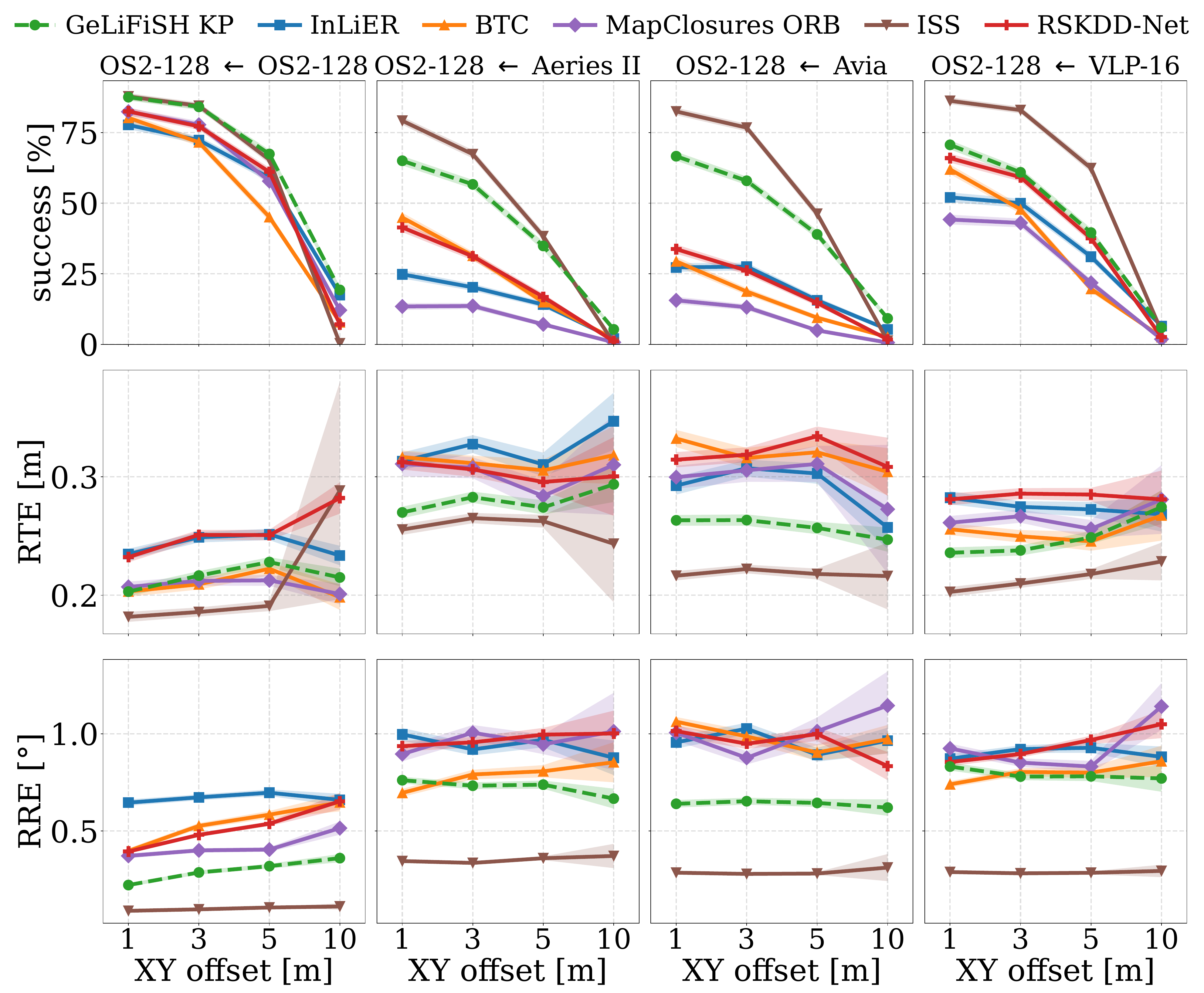}
    \caption{Keypoint-only registration on the pairs of Fig.~\ref{fig:repeatability}, by XY offset. Registration recall is the fraction of pairs with RTE below 0.5\,m and RRE below $2^\circ$, and RTE and RRE are averaged over the successfully registered pairs. Shading is $\pm1$ standard deviation over a bootstrap of the keyframes.}
    \label{fig:registration}
    \vspace{-12pt}
\end{figure}

For registration (Fig.~\ref{fig:registration}), the number of keypoints matters as much as their repeatability.
Although as repeatable as GeLiFiSH, BTC retains too few keypoints to constrain ICP across sensors, and at 1\,m reaches a registration recall of 29.4 to 62.0\,\% on the cross-sensor pairs, against 65.0 to 70.7\,\% for GeLiFiSH.
ISS, with the most keypoints, achieves the highest recall at 1 and 3\,m on every sensor, with GeLiFiSH second in each case and on par with ISS on the Ouster.
At 5\,m, GeLiFiSH surpasses ISS on the Ouster.
Compared to InLiER, GeLiFiSH increases the recall at 1\,m by 40.2 and 39.4 points on the Aeva and the Avia.
Over the successfully registered pairs up to 5\,m, its RTE and RRE are also lower, 0.25\,m and $0.62^\circ$ against 0.29\,m and $0.88^\circ$.
At 10\,m, no method exceeds 20\,\%, as ICP initialized at the identity rarely converges over such offsets~\cite{stathoulopoulos2026inlier}.

The full pipeline processes a scan in 22.4 (Avia) to 56.7\,ms (Ouster) on a single core, including GS, MOS, and RKE, below the 100\,ms period of the 10\,Hz sensors, and the delayed labels add 1.0 to 7.4\,ms.
RKE accounts for 5.3 to 17.2\,ms per scan, or 83 to 172\,ms per window, within the 6.8 (MapClosures) to 207\,ms (ISS) that the baselines require per window, which they can only begin once all scans of the window are recorded.
Overall, GeLiFiSH is as repeatable as the best baseline while providing several times its keypoints, achieves the second-highest registration recall after ISS with a fraction of its keypoints, and updates its keypoints after every scan, with the ground and moving labels included in its runtime.

\section{Conclusions\label{sec:conclusion}}
We presented GeLiFiSH, a learning-free filter that recursively estimates persistent LiDAR keypoints at scan rate.
Detections on a rolling map are maintained as tracks that a sequential probability ratio test accepts or rejects, ground segmentation provides a sensor-agnostic height reference, and moving object segmentation removes dynamic points before map insertion.
With a single parameter set across five LiDARs, GeLiFiSH attains the highest average chance-corrected repeatability across sessions and sensors, consistently exceeds the same detector applied to submaps, and achieves the second-highest registration recall from keypoints alone, behind a detector with about four times as many keypoints.
Without a sensor-height prior, its ground segmentation exceeds Patchwork++ on SemanticKITTI and agrees more closely across the HeLiMOS sensors, its moving object segmentation attains the highest mean IoU over these sensors, and the full pipeline runs within the scan period on a single CPU core.
Future work will exploit the keypoint covariances in multi-session registration and evaluate them within place recognition and loop closure.





\bibliographystyle{IEEEtran}
\bibliography{IEEEabrv,references}

\end{document}